\documentclass{article}

\usepackage{arxiv}

\usepackage[utf8]{inputenc} 
\usepackage[T1]{fontenc}    
\usepackage{hyperref}       
\usepackage{url}            
\usepackage{booktabs}       
\usepackage{amsfonts}       
\usepackage{nicefrac}       
\usepackage{microtype}      
\usepackage{lipsum}		
\usepackage{graphicx}
\usepackage{natbib}
\usepackage{doi}
\usepackage{import}
\usepackage{flafter}
\usepackage{placeins}
\usepackage{subcaption}

\title{
Share, Collaborate, Benchmark: Advancing Travel Demand Research through rigorous open-source collaboration.}

\author{ \href{https://orcid.org/0000-0003-0560-6223}{\includegraphics[scale=0.06]{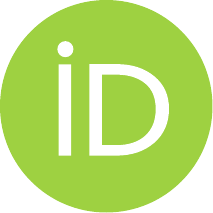}\hspace{1mm}Juan D.~Caicedo}\thanks{Coresponding Author} \\
	Department of Civil and Environmental Engineering\\
    Global Metropolitan Studies \\
	University of California, Berkeley, USA\\
	\texttt{juan\_caicedo@berkeley.edu} \\
    \And
	\href{https://orcid.org/0009-0001-7713-6171}{\includegraphics[scale=0.06]{orcid.pdf}\hspace{1mm}Carlos Guirado} \\
	Department of Civil and Environmental Engineering\\
	University of California, Berkeley, USA\\
	\texttt{guirado@berkeley.edu} \\
	\And
	\href{https://orcid.org/0000-0002-8482-0318}{\includegraphics[scale=0.06]{orcid.pdf}\hspace{1mm}Marta C.~Gonz\'{a}lez} \\
	Department of Civil and Environmental Engineering\\
    Department of City and Regional Planning \\
	University of California, Berkeley, USA\\
	\texttt{martag@berkeley.edu} \\
	\And
    \href{https://orcid.org/0000-0002-4407-0823}
    {\includegraphics[scale=0.06]
    {orcid.pdf}\hspace{1mm}Joan L.~Walker} \\
	Department of Civil and Environmental Engineering\\
    Global Metropolitan Studies \\
	University of California, Berkeley, USA\\
	\texttt{joanwalker@berkeley.edu} \\
}

\date{}

\renewcommand{\headeright}{}
\renewcommand{\undertitle}{}
\renewcommand{\shorttitle}{Share, Collaborate, Benchmark}

\hypersetup{
pdftitle={second_paper},
pdfsubject={meta-analysis},
pdfauthor={Caicedo, J. Gonzalez, M. Walker, J},
pdfkeywords={Open-Source Benchmarking, Short-Term Ridership Prediction, Highly Dynamic Conditions}}

\begin{document}
\maketitle
\begin{abstract}
This research foregrounds general practices in travel demand research, emphasizing the need to change our ways. A critical barrier preventing travel demand literature from effectively informing policy is the volume of publications without clear, consolidated benchmarks, making it difficult for researchers and policymakers to gather insights and use models to guide decision-making. By emphasizing reproducibility and open collaboration, we aim to enhance the reliability and policy relevance of travel demand research. We present a collaborative infrastructure for transit demand prediction models, focusing on their performance during highly dynamic conditions like the COVID-19 pandemic. Drawing from over 300 published papers, we develop an open-source infrastructure with five common methodologies and assess their performance under stable and dynamic conditions. We found that the prediction error for the LSTM deep learning approach stabilized at a mean arctangent absolute percentage error (MAAPE) of about 0.12 within 1.5 months, whereas other models continued to exhibit higher error rates even a year into the pandemic. If research practices had prioritized reproducibility before the COVID-19 pandemic, transit agencies would have had clearer guidance on the best forecasting methods and quickly identified those best suited for pandemic conditions to inform operations in response to changes in transit demand. The aim of this open-source codebase is to lower the barrier for other researchers to replicate, reproduce models and build upon findings. We encourage researchers to test their own modeling approaches on this benchmarking platform, challenge the analyses conducted in this paper, and develop model specifications that can outperform those evaluated here. Further, collaborative research approaches must be expanded across travel demand modeling if we wish to impact policy and planning.
\end{abstract}


\newpage

\section{Introduction}

In an era of rapid technological change, shifting social norms, and disruptive events, the travel demand research field finds itself increasingly unable to match the needs of the reality it aims to model with its traditional methodologies and research practices. An example that highlights this disconnect is the challenge of demand forecasting under highly dynamic conditions characterized by abrupt changes, a scenario that is increasingly common. This mismatch between the field's capabilities and the fluctuating real-world context it seeks to understand limits its ability to provide timely, actionable insights and policy-relevant decision support to practitioners such as transit agencies and policymakers navigating these disruptions.

A core challenge is the lack of emphasis on reproducibility and systematic comparability across research practices. While many papers provide methodological details, the underlying code implementation is rarely shared, leaving other researchers to potentially re-implement methods inconsistently. Dataset sharing is also uncommon, making full replication nearly impossible. The proliferation of studies further exacerbates this issue: with so many proposed techniques, benchmarking efforts can only include a small subset due to resource constraints, leading to fragmented and incomplete comparisons. This obstructs the development of collective wisdom, since it becomes difficult to identify the most promising approaches and assess the robustness of findings. The traditional incentives and publication process in travel demand research prioritize claims of novelty and uniqueness over ensuring reproducibility, replicability and comparability. This is apparent in other fields as well \citep{nosek2012ii}.

As highlighted by \citep{munafo2017}, across scientific disciplines there is substantial empirical evidence indicating threats to knowledge accumulation, such as low statistical power and lack of data sharing. These issues can dramatically increase the probability that published findings are incorrect or cannot be reproduced or applied elsewhere \citep{munafo2017,ioannidis2005, baker2016}. Reproducibility and replicability are central to providing confidence in findings \citep{nosek2012ii} and hence are vital for research to effectively inform policy and decision-making, which should be the core mission of the travel behavior literature. Without the ability to verify, scrutinize and build upon findings, policy relevance and potential impact is severely diminished \citep{nosek2012s}. This has undermined public trust and the role of science in evidence-based policymaking \citep{hendriks2020,sanders2018,korbmacher2023}. Recognizing this, fields like machine learning have embraced open, collaborative and reproducible practices, which has catalysed collective progress. The NeurIPS 2019 reproducibility program shows how mechanisms like code submission policies, community challenges and reporting checklists can improve scientific standards and assess the reliability of findings to better enable real-world applications \citep{pineau2021}. Other fields have also transitioned towards more open, reproducible research practices to rebuild credibility, as exemplified by the systematic changes resulting from the "credibility revolution" in psychology and other behavioral and social sciences \citep{korbmacher2023}.

The focus of this paper on short-term transit demand forecasting under highly dynamic conditions is just one application area within the broader travel demand research field. It serves as an illustrative example, but the principles advocated here of building open, shared research infrastructure for enhanced reproducibility can and should be applied across other domains of our literature as well: be it long-term travel demand forecasting, activity-based models, behavioral insights, or other core areas, embracing open science practices like collective data/code sharing, systematic replication , and collaborative benchmarking can accelerate the field's ability to self-evaluate, identify where the state of the art lies, and translate findings into policy in order to keep up with our rapidly changing world.

One of the key obstacles hindering the travel behavior research literature from having a substantive impact on policy is the overwhelming number of publications lacking well-defined and widely-accepted benchmarks. This makes it challenging for policymakers to extract actionable insights from the vast literature. Researchers also face this difficulty: in preparing this paper, we reviewed over 300 published works studying transit ridership prediction. However, we found it virtually impossible to conclusively synthesize where the state-of-the-art stands across these diverse areas of research. The lack of systematic replication, consolidated benchmarking exercises, and open sharing of fully-fledged methods, data, and code impedes our capacity to identify which techniques are reliable, policy-relevant and ready for deployment. Without mechanisms to effectively evaluate the strengths and limitations of different proposed approaches, or to validate findings across a range of real-world conditions, policy uptake will remain extremely limited. Instead, we are facing a growing literature focused solely on claims of novelty and uniqueness, which cannot substantially effect policy. For travel behavior research to impact transportation policy and planning in practical ways, we must transition towards a more open, reproducible, and self-correcting research ecosystem.

The COVID-19 pandemic starkly highlighted these limitations. As transit agencies urgently required data-driven guidance to adapt operations, existing predictive models struggled to keep pace with the sudden, highly dynamic shifts in demand patterns. Some agencies made adjustments uniformly across cities, failing to account for how certain communities may have greater transit needs, and how these needs evolved over time. This underscores the necessity for agile research practices that can rapidly validate the performance of models under changing conditions and provide reliable decision support.

To address these challenges, this paper proposes a fundamental reorientation of travel demand research through more collaborative and open approaches. Specifically, we advocate for the development of an open-source platform to facilitate consistent benchmarking, comparative analysis and collective refinement of predictive models across diverse scenarios.

The key contributions of this paper are twofold: First, we extend an open invitation to the travel demand research community to actively engage with and build upon our open-source codebase and dataset. Our infrastructure provides researchers access to a centralized pool of validated models, streamlining implementation and evaluation against consistent standards. Moving towards collaborative and open research practices will allow the community to produce reliable and relevant insights that can inform rapid decision-making in practical policy contexts. Second, we leverage this infrastructure to conduct rigorous comparative analyses that evaluate the strengths, weaknesses and adaptability of different model types under stable and highly dynamic conditions like the effect of the COVID-19 pandemic using 5-years data from the BRT system in Bogot\'{a}, Colombia.

The remainder of the paper is organized as follows. Section 2 includes an overview of multiple model architectures in the field of transit demand ridership, focusing on a few these architectures and two modeling strategies for short-term public transit demand forecasting. Next, Section 3 demonstrates how an open-source implementation of transit ridership prediction models can address policy questions that current travel demand research practices cannot answer promptly. Then, Section 4 presents the meta-analysis results. This section emphasizes the importance of comparison and suggests that modeling strategies, largely independent of specific methods, might play a more crucial role than the model architecture itself. Lastly, Section 5 offers a discussion and a call to action to reorient research practices towards emphasizing openness, reproducibility and comparability to build collective wisdom.

\section{Literature Review}

The analysis focuses on the short-term public transit demand forecasting problem, which has seen a proliferation of proposed techniques spanning statistical models \citep{Milenkovic2018, Ding2017, Sun2014}, machine learning \citep{Sun2015, Li2017, Ding2016, Toque2018, Dai2018,  Chen2011}, deep learning \citep{Tsai2009,Wei2012,9141203,Liu2019,8951085,Borovykh2017ConditionalTS,cho-etal-2014-properties,6795963, ijgi8060243, ZHANG2021102928, Bai2017}, Bayesian approaches \citep{Roos2017}, optimization \citep{Glisovic2016, Xie2020, Yuan2019}, and even a mixture of approaches \citep{Jing2021,MA2014148,Zhang2022,Zhang2019}. However, due to lack of reproducibility and consolidated benchmarking, there has been little systematic study of how these models fare in disrupted scenarios with abrupt demand shifts.

While the literature on transit ridership prediction suggests numerous approaches, this section focuses on the five most prevalent methods. Beyond these methods, we spotlight two crucial yet often ignored modeling designs —single vs. multi-output and static vs. online training strategies—which are mostly independent of the model, yet scarcely mentioned in the literature on transit ridership prediction. For highly dynamic conditions, we also reference literature on demand prediction for special events, such as holidays, concerts, and sporting events. In addition, we primarily focus on aggregate prediction, which aggregates demand at the station or bus-route level.

Parametric models, such as ARIMA and SARIMA, have traditionally been used to model time-series data. The aim is to model a stationary process where the mean is zero and the variance is constant. The models comprise autoregressive and moving average parts to capture historical data and past prediction errors, and SARIMA accounts for trends and seasonal patterns. The main advantages of these models are the solid statistical background and default calculation of prediction confidence intervals. Additionally, \cite{Engle1982} and \cite{Bollerslev1986} proposed an autoregressive conditional heteroskedasticity (ARCH) and generalized autoregressive conditional heteroskedasticity (GARCH) model to relax the constant variance assumption of the stationary process, allowing the modeling of heteroskedasticity in time-series analysis. \cite{Milenkovic2018} used a SARIMA model to forecast monthly demand using 10 years of training data. Moreover, \cite{Ding2017} used an ARIMA with GARCH model to model demand volatility, using 15-min aggregation for one month of data and three transit stations with high passenger demand. 

Deep learning models have also been widely used for time-series analyses in the last decade. There are three main deep learning architectures: The MLP\citep{Tsai2009,Wei2012}, the CNN \citep{9141203}, and the RNN \citep{Liu2019,8951085}. The MLP model is a fully connected feed-forward neural network with at least three layers: the input, hidden, and output layers. For a time series, the input layer comprises past observations; however, given that it is a fully connected network, there is no explicit time dependency. Inspired by image classification, the CNN model connects inputs with predefined temporal dependencies (e.g., observations within the same week) \citep{Borovykh2017ConditionalTS}. The RNN allows the outputs of the previous nodes to influence subsequent nodes, which is a more realistic representation of time. In the RNN, the most common cells are the LSTM \citep{6795963} and gated recurrent unit (GRU) \citep{cho-etal-2014-properties}, which maintain hidden states to filter relevant information and store long-term dependencies. 

There are two main model designs to predict short-term ridership. Single- versus multi-output models and the adaptive training strategy. Single-output models train individual target values (e.g., transit stations), and multi-output models train one model to predict the multiple interdependent target variables from a given set of input variables (e.g., multiple stations in the transit system). In the parametric case, they are known as vector autoregressive models (VAR) \citep{Kirchgassner2007}. Multi-output models have been used in other fields, such as the energy forecast prediction \citep{Sajjad2020} and the air quality prediction \citep{Zhou2019}. The main advantages of these models are that they capture the spatial and temporal correlation, and the modeler needs only to train one model instead of multiple individual models for each time series. In transportation research, multi-output models have recently been used for bike-sharing system demand prediction \citep{s22031060}, bus travel time prediction \citep{Petersen2019}, and public transit passenger prediction \citep{Wang2022}. 

In addition, static models maintain the estimated parameters for prediction, whereas adaptive training models (online or continuous learning models) modify the parameters when new information becomes available. This paper refers to adaptive training as online training. The idea of online training comes from the problem that artificial neural networks forget past information when trained for a new task, known as catastrophic forgetting \citep{HADSELL20201028}. In practice, ARIMA and SARIMA models update the model parameters when new information is available, as suggested by the default settings of multiple implementations \citep{pmdarima,Loning2019}. However, this aspect is not underscored in machine learning and deep learning models for short-term ridership prediction. In machine learning, this modeling strategy has been explored in other fields, such as energy consumption \citep{AHMAD2020102010}, medical research \citep{LEE2020e279}, recommendation systems \citep{PORTUGAL2018205}, and spam detection \citep{doi:10.1080/1206212X.2020.1751387}, but it has not been applied to the short-term demand prediction problem of public transit. One of the advantages of online training is its adaptability to unpredictable and uncertain changes and the ability of these models to support real-time decision-making, which is relevant in highly dynamic conditions. 

Primarily, previous methods have focused on short-term demand prediction under stable conditions. However, demand prediction during significant events has garnered attention. For example, \cite{WANG2019580} developed an early warning system that detects abnormal passenger outflows to predict abnormal passenger inflows in the future and cross-referenced their finding with one large-scale event. Similarly, \cite{doi:10.1177/03611981211047835} proposed a clustering mechanism and dynamic time wrapping to account for multiple social events such, as holidays and sporting events. Additionaly, \cite{doi:10.1177/03611981221086645} implemented a naïve Bayesian-based transition model to forecast demand under unplanned events. These methods, however, have two main shortcomings. First, they were only validated during known periods of disruption, such as holidays and large-scale events, but not under highly dynamic conditions. Second, these methods do not take full advantage of multi-output models, as they only consider specific stations where special events are likely to happen.
\section{Methods and Data}

One of the key research objectives is to develop an open-source implementation of several transit ridership prediction models in the literature. Instead of fixating on the methods themselves (details of which can be found in the original papers), we critique the general framework, examining how these methods perform under various modeling strategies in highly dynamic conditions. Our goal is not to propose a new method, but to demonstrate the significant improvements in research practices that could be achieved with open-source implementations that emphasise reproducibility and comparability.

\subsection{Method Selection}

Table \ref{tab:table1} includes at least one reference for each major methodology mentioned in the literature, which has been used to inspire the implementation of the open-source codebase. We prioritized papers with high citation counts, clear model explanations, and relatively canonical model implementations for each category as a first step to build the open-source platform. This approach ensured building on well-established and rigorously tested methodologies with proven effectiveness for short-term ridership prediction. To ensure the comparability of the models, we applied the same preprocessing and feature engineering techniques across all implementations, even if it meant deviating from the exact methodology described in some of the selected papers. This approach was necessary to ensure that the methods were appropriately adapted to the specific data while maintaining consistency in the evaluation process. For instance, \cite{Liu2019} implemented three LSTM networks to model same-day, daily, and weekly demand. In the proposed implementation, demand is aggregated daily, so we use only one LSTM network. While some studies have found that external data, such as weather, land use, and social data, improve accuracy, we avoided using them because this research objective is not to test the accuracy of external data but the performance of the methods themselves. 

\begin{table}[ht]
\small
\caption[Short-Term Ridership Prediction Selected References]{Short-Term Ridership Prediction Selected References for Open-Source Codebase Infrastructure Implementation (Attributes as Reported in the References) }
\centering
\begin{tabular}{llllllll}
    \toprule[1.5pt]
    Study     & Methods    & \shortstack{Number of \\ Stations} & Time Frame & Aggregation & Accuracy & Preprocessing \\
    \midrule[1.5pt]
    \cite{Ding2017}  & ARIMA & 3 & \shortstack{1 Month\\October 2012} & 15 mins & \shortstack{MAPE: \\ 3.39\%} &  \shortstack{Log \\ Transformation}   \\
    \cmidrule(r){2-7}
    \cite{Milenkovic2018} & SARIMA & Full System & \shortstack{10 years \\ 2004-2014} & Monthly & \shortstack{MAPE: \\ 7.13\%} &  \shortstack{Log \\ Transformation}   \\
    \cmidrule(r){2-7}
    \cite{Wei2012} & MLP & 1 Line & \shortstack{1 Month\\May 2008} & 15 mins & \shortstack{MAPE: \\ 5.04\%} & \shortstack{Empirical Mode\\ Decomposition}   \\
    \cmidrule(r){2-7}
    \cite{9141203} & CNN & 1091 Lines & \shortstack{8 Months \\ Mar-Oct 2016} & Hourly & \shortstack{SMAPE: \\ 20.98\%} & Unclear    \\
    \cmidrule(r){2-7}
    \cite{Liu2019}  & LSTM & 3 & \shortstack{5 Months}  & 10 mins &  \shortstack{SMAPE \\(DL-N)\\ 16.91\%} & Min-Max \\
    \bottomrule[1.5pt]
\end{tabular}
\label{tab:table1}
\end{table}

\cite{Milenkovic2018} used the SARIMA model to account for annual seasonality patterns. \cite{Ding2017} proposed a  GARCH with ARIMA model to account for demand volatility. For this research, we only trained the model with the ARIMA part because the GARCH component only affects the estimation of the demand volatility but not the point estimate. These two research selected a model based on an information criterion. \cite{Ding2017} used the Akaike information criterion (AIC) score and \cite{Milenkovic2018} used the Bayesian information criterion (BIC) score for some predefined models. Model selection is a time-consuming task, especially when training a model for an individual station; thus, we used a stepwise algorithm by \cite{JSSv027i03} to expedite the model selection using the AIC score. 

\cite{Wei2012} used a three-layer neural network where the input layer contains as many nodes as input variables, and the output layer size is the forecast window. The hidden layer size was taken from the default implementation of NeuroShell 2, which is the average of the input and output layer sizes. Our implementation of the MLP model uses the same model structure. \cite{Liu2019} proposed many data sources and feature engineering procedures. However, for this research, we selected the implementation of their deep learning nearest part architecture (DL-N) model, which only takes the information from previous timesteps. However, as this research focuses on daily demand forecasting based on a full week's demand, this implementation considers daily and weekly cycles. The original implementation used a unique LSTM layer with 32 units, which is maintained in our implementation. The CNN architecture consists of one convolutional layer with 256 filters and a dilatation factor of seven to represent the weekly cycle. 

\subsection{Experiments}

To compare the model implementations in \ref{tab:table1}, we conducted four sets of experiments varying the model design in two critical dimensions that can affect model performance under highly dynamic conditions. The first dimension considers two learning strategies: static and online. The static learning strategy estimates a model using only the test data and generates predictions using the latest available information, but the model is not re-trained with new data. In the online approach, the model is re-trained as new information becomes available. We compared a single-output to a multi-output model in the second design dimension. With the single-output strategy, we trained each transit station's time-series data individually. In the multi-output strategy, we used the time-series data from all stations as input to predict their demand simultaneously.

To train the ARIMA and SARIMA models, we used the Python implementation by \cite{pmdarima}. However, the implementation is restricted to univariate time-series, and requires updating model parameters with new information. Given these constraints, we only tested the ARIMA and SARIMA models for the single-output and online-training experiments. Table \ref{tab:table2} summarizes the set of experiments.

\begin{table}[h!]
	\caption{Modeling Strategies}
	\centering
	\begin{tabular}{c|cc}
		\toprule[1.5pt]
		Strategy    &  \shortstack{Single \\ Output}   & \shortstack{Multi-output}\\
		\midrule[1.5pt]
		\shortstack{Static \\ Training}  & \shortstack{MLP\\CNN\\LSTM} & \shortstack{MLP\\CNN\\LSTM} \\
        \cmidrule(r){2-3}
        \shortstack{Online \\ Training}  & \shortstack{MLP\\CNN\\LSTM\\ARIMA\\SARIMA} & \shortstack{MLP\\CNN\\LSTM} \\
		\bottomrule[1.5pt]
	\end{tabular}
	\label{tab:table2}
\end{table}

\subsection{Data}

\begin{figure}[h!]
\begin{center}
\includegraphics[width=0.95\textwidth]{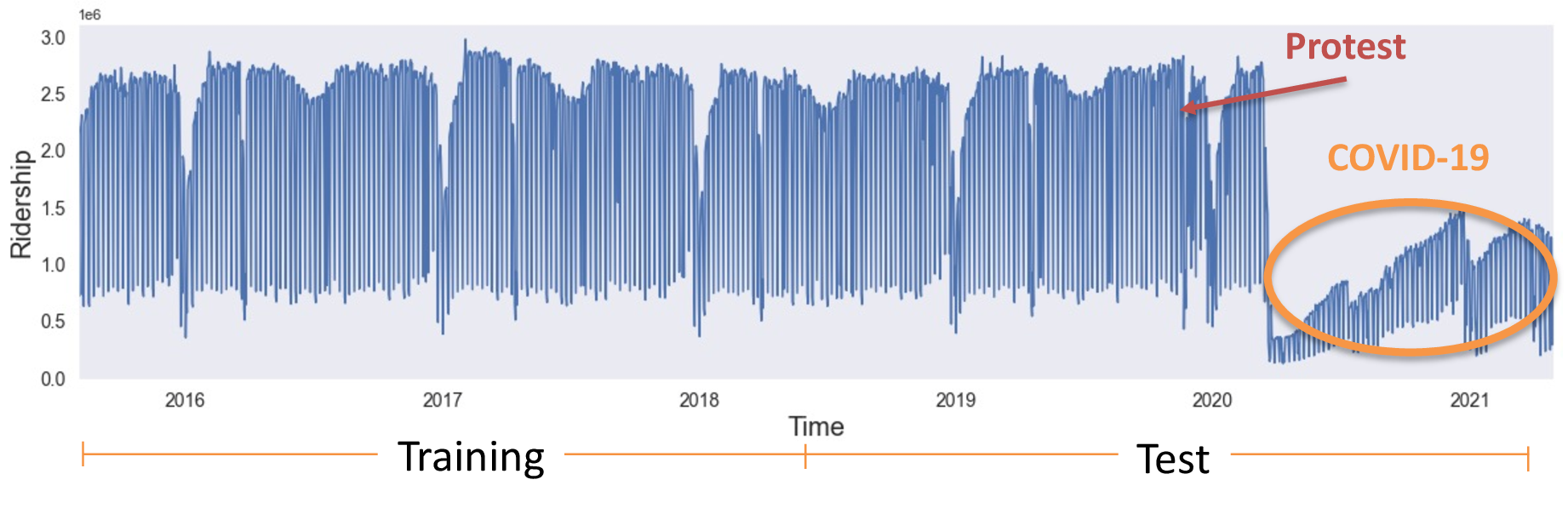}
\end{center}
\caption{BRT Daily Aggregated Demand From August 2015 to May 2021. Training Period: August 2015 to July 2018. Test Period: August 2018 to May 2021. Protest: November and December 2019. COVID-19: March 2020 to May 2021}\label{fig:fig1}
\end{figure}

The data encompass the daily transactions of 147 stations over five years (August 2015 to May 2021) in the BRT system of Bogot\'{a}, Colombia, as presented in Figure \ref{fig:fig1}. The BRT system is the city's primary mass public transportation system, with over 2.5 million daily transactions during stable conditions (pre-COVID-19). There were two highly dynamic conditions during this period—a major national protest in November and December 2019 and the ongoing COVID-19 pandemic since March 2020. Both events triggered unexpected and uncertain closures of multiple transit stations. This dataset is publicly available at: https://datosabiertos-transmilenio.hub.arcgis.com. It is worth noting that this dataset is updated daily. The inability to answer basic questions, such as identifying the best-performing model during these conditions, stems not from a lack of data but from the lack of necessary tools and collaborative frameworks to conduct efficient analysis.

We trained a baseline model with data from April 2015 to August 2018 for the four sets of experiments. The testing period starts in September 2018 and ends in April 2021. Next, we used the baseline mode for the online-training strategy and updated it at every timestep. Further, we normalized the data for all experiments using the min-max approach because it normalizes element values that fall between zero and one and maintains the meaning of zero (zero transactions). The temporal resolution of the data was deliberately set at one day, taking into consideration that finer resolutions would result in a significantly larger dataset, particularly when working with a five-year-long dataset. This choice was motivated by the limitations of ARIMA and SARIMA models, which do not scale well with larger data sets. Consequently, due to computational constraints, it was not feasible to fit these models with a finer resolution. As suggested by \cite{Liu2019}, we used the past information of the last three weeks to predict the daily demand for the following week. We included standard temporal variables to account for cyclical patterns. We incorporated dummy variables to capture the distinctive demand behavior observed on Saturdays and holidays (where holidays encompass both Sundays and other holidays). Additionally, to encode the yearly and weekly patterns, we employed sine and cosine functions.

\subsection{Analysis Metric}
The most common performance indicators used in the literature for short-term demand forecasting are the root mean squared error (RMSE), mean absolute error (MAE), and mean absolute percentual error (MAPE). The RMSE and MAE measures depend on the magnitude of the observations, making them helpful in comparing different methods on the same dataset but not as effective when comparing one method across multiple datasets. The MAPE solves that problem by estimating a percentage error instead of an error. In the MAPE, the error can be considered a slope of $\widehat{|y}-y|/|y|$, thus bounding the error within the range 0 to +$\infty$. However, this metric might result in large error as the target value $y$ approximates zero. Such situations frequently occur when stations are closed or no transactions are recorded within a certain time frame. Disregarding this information is not advisable, especially in the case of multi-output models, where station closures can provide valuable insights for those stations still operational. In highly dynamic conditions, station closures or service disruptions are more prevalent due to external forces, thus making the MAPE an inadequate metric of analysis. 

Instead, we use the mean arctangent absolute percentage error (MAAPE)\citep{Kim2016}, which limits the bounds of the error by considering the error to be an angle. The MAAPE applies the arctanget function to the absolute percentage error, limiting the bounds of the error from 0 to $\pi/2$, making this metric more robust to outliers. The advantages of this metric is that it preserve the advantages of the MAPE, such as the scale-free aspect, and can also be interpreted as an absolute percentage error. We do not advocate for the MAAPE metric as the sole metric of analysis, nor do we assert that it should be the gold standard for transit ridership prediction models. We use this metric due to its ability to capture information when zero transactions are recorded at a station, accurately reflecting the system's true behavior in highly dynamic conditions. Ultimately, the selection of the performance metric should align with the model's specific objectives.

This research uses a system-wide level that tests the performance of the entire transit system at every time step. Unlike the current trends in the literature that test accuracy at the individual level, this metric serves as a summary metric to test the overall performance of the model. The advantage of having a system-wide metric is that it captures the performance of a broader range of transit behaviors, making the metric more generalizable. Therefore, it can also compare performances in multiple geographical areas. The average error of the entire transportation system is given as follows:  


\begin{equation}
	{MAAPE}_t=\frac{1}{s\ast w}\sum_{j=1}^{s}\sum_{i=1}^{w}{\arctan\left(\frac{{\hat{y}}_{s,w,t}-y_{s,w,t}}{y_{s,t,w}}\right)}
\end{equation}

where $t$ denotes the aggregation period, $s$ represents the number of stations, and $w$ indicates the forecast window. Therefore, $\hat{y}_{s,w,t}$ represents the predicted values of station $s$ for $w$  forecast windows in the period $t$. 
\section{Results}
This section presents the empirical findings obtained from the conducted experiments, encompassing the system-level outcomes and model performance in both stable and highly dynamic conditions. We first show the chronological evolution of the MAAPE for all experiments, followed by a comparative analysis of the model performance across the different conditions examined in this research.

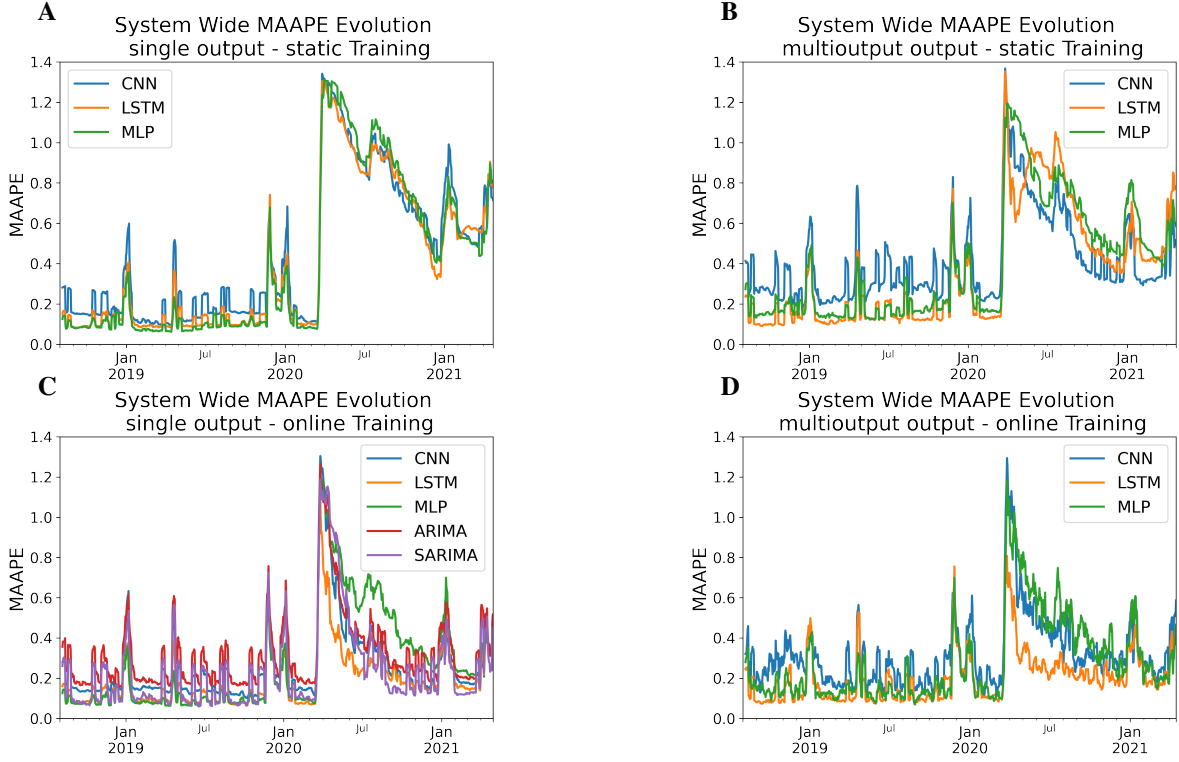
\begin{figure}[ht]
  \centering
  \vspace{0.5cm}
  \begin{minipage}{0.45\textwidth}
    \centering
    \begin{subfigure}{\textwidth}
      \begin{picture}(0,0)
        \put(20,140){\textbf{A}}
      \end{picture}
      \includegraphics[width=\textwidth]{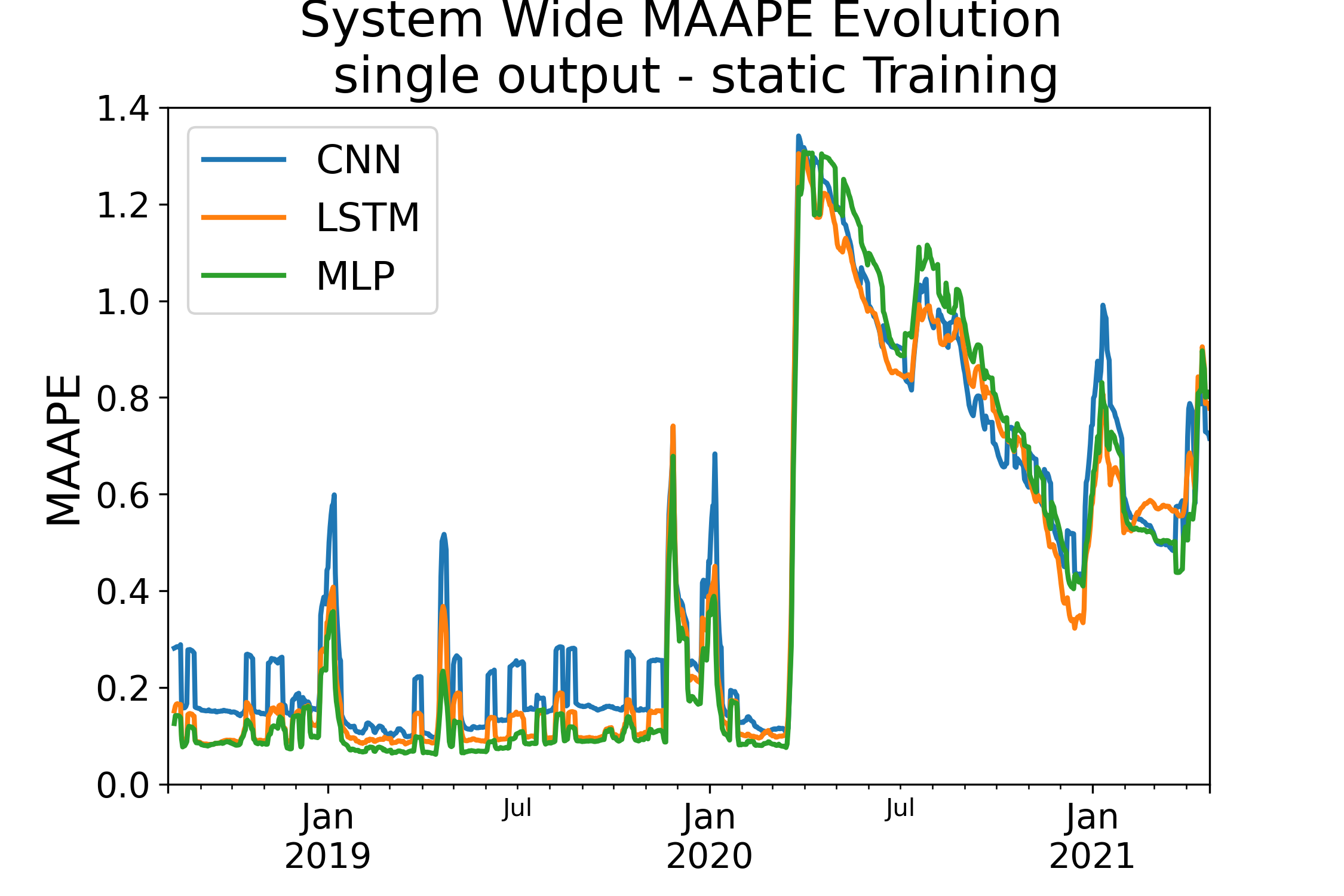}
    \end{subfigure}
    \vfill
    \begin{subfigure}{\textwidth}
      \begin{picture}(0,0)
        \put(20,140){\textbf{C}}
      \end{picture}
      \includegraphics[width=\textwidth]{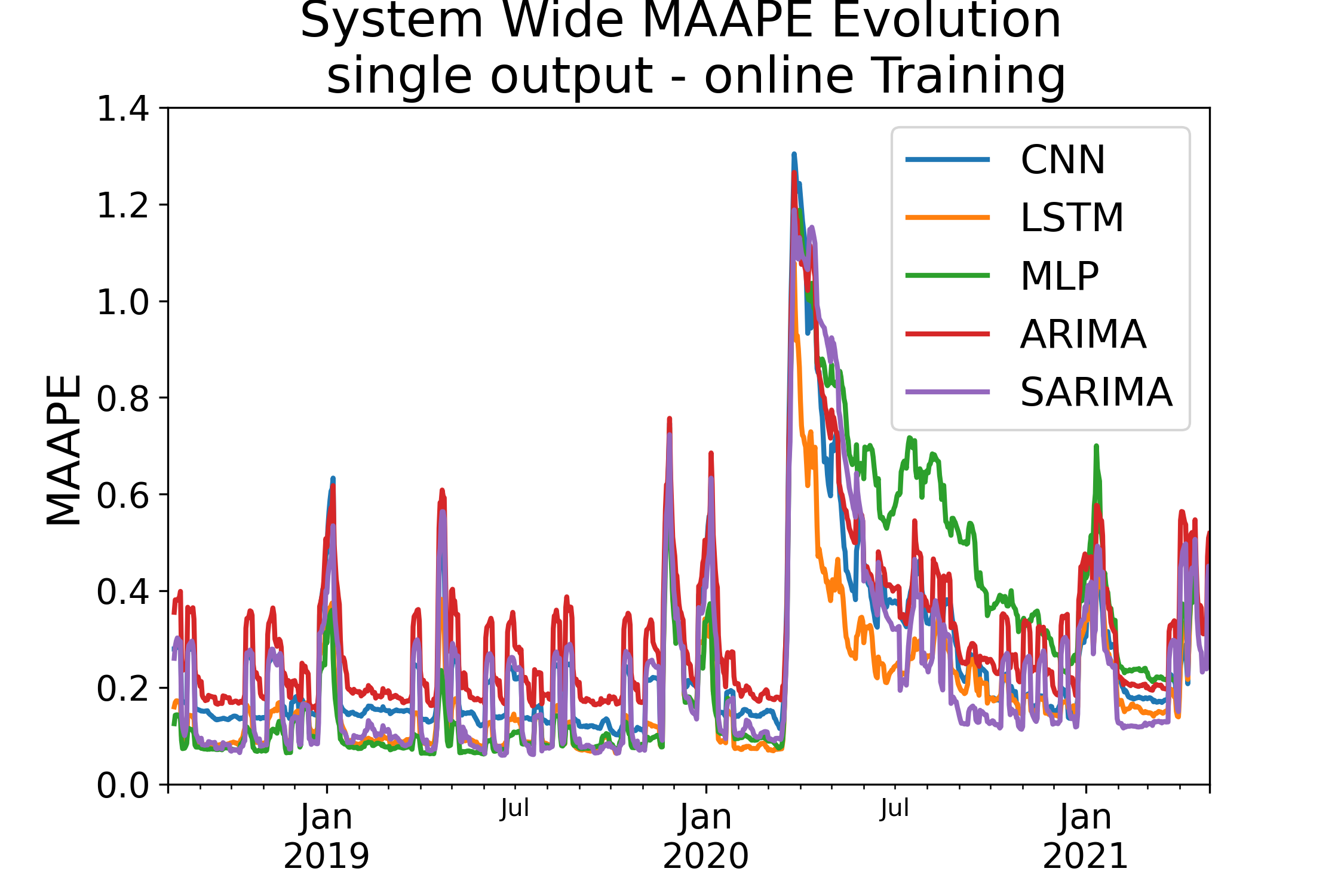}
    \end{subfigure}
  \end{minipage}
  \hfill
  \begin{minipage}{0.45\textwidth}
    \centering
    \begin{subfigure}{\textwidth}
      \begin{picture}(0,0)
        \put(20,140){\textbf{B}}
      \end{picture}
      \includegraphics[width=\textwidth]{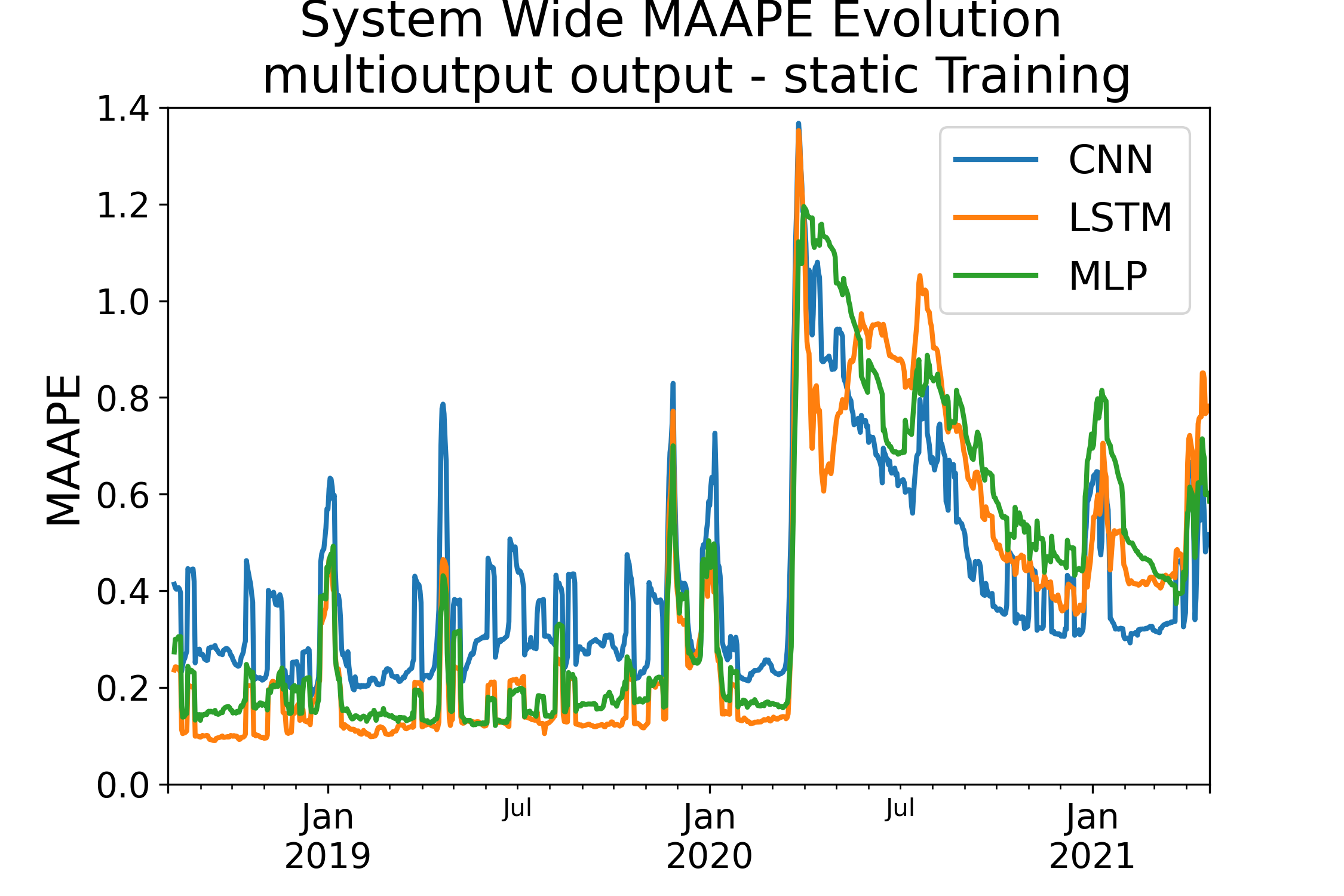}
    \end{subfigure}
    \vfill
    \begin{subfigure}{\textwidth}
      \begin{picture}(0,0)
        \put(20,140){\textbf{D}}
      \end{picture}
      \includegraphics[width=\textwidth]{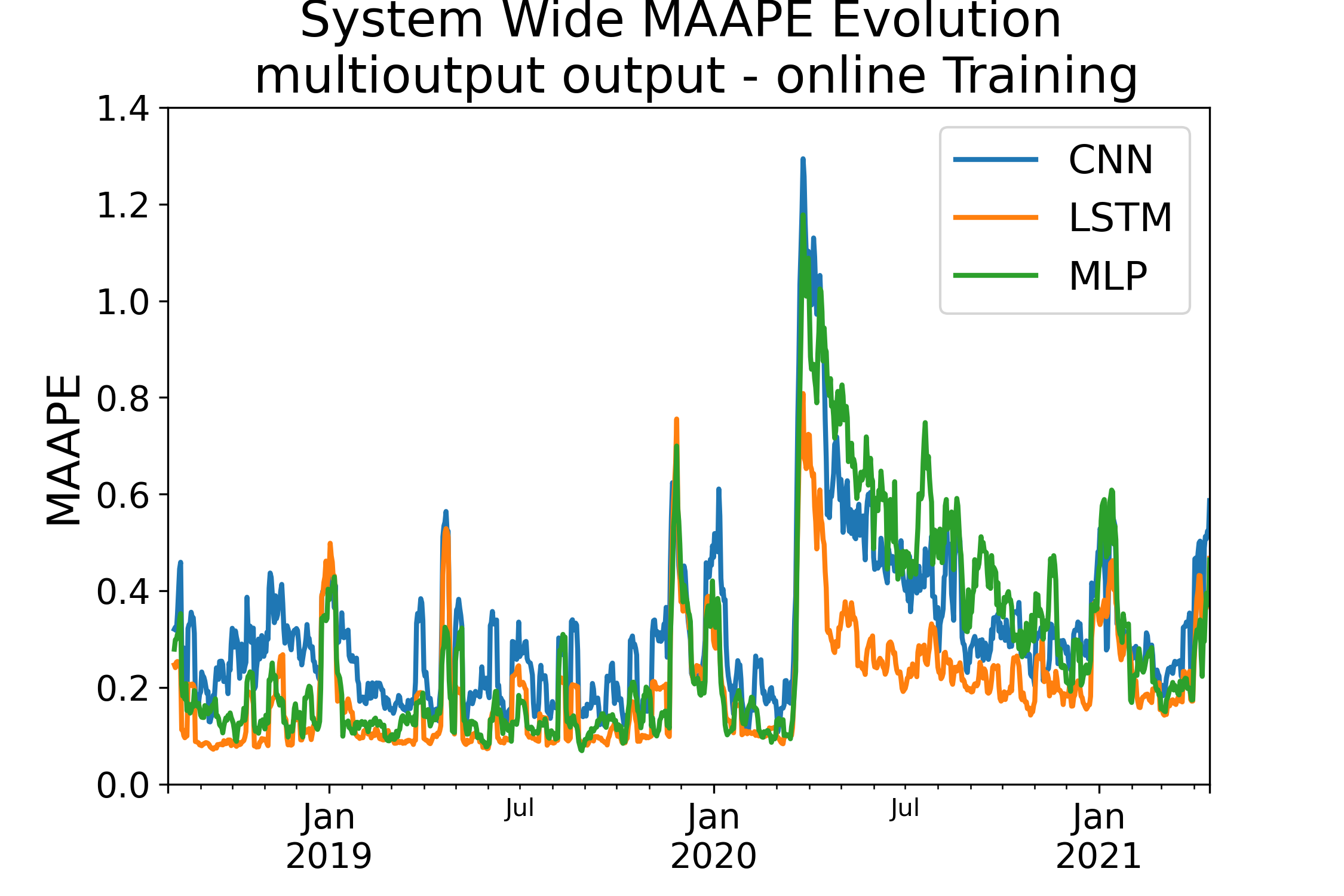}
    \end{subfigure}
  \end{minipage}
  \caption[Daily System-Wide Mean Arctangent Absolute Percentage Error Evolution for the Testing Period ]{Daily System-Wide Mean Arctangent Absolute Percentage Error Evolution for the Testing Period . (A) Single-output and static training, (B) Multi-output and static training, (C) Single-output and online training, and (D) Multi-output and online training.}
  \label{fig-2-2:maape_evolution}
\end{figure}

To understand the evolution of the error metric, we estimated a daily system-wide MAAPE for all experiments, as depicted in Figure \ref{fig-2-2:maape_evolution}. The plot is smoothed using the rolling MAAPE for the last seven days. Most models performed similarly in all four experiments in stable conditions, although the CNN and ARIMA models consistently underperformed. During stable conditions, some peaks in the MAAPE metric correspond to the end-of-year holidays, Eastern holidays, and long weekends. These results also reveal a notable trend across all models and experiments, characterized by a significant surge in the MAAPE during the two highly dynamic conditions. However, it is important to highlight that the initial impact observed in the month-long protest condition was relatively milder compared to the COVID-19 pandemic. The online-training experiments revealed a faster decrease in the MAAPE during the COVID-19 pandemic; however, there is a noticeable difference in the expected time for the MAAPE to stabilize. On average, models in the single-output and online-training experiments take three months, whereas the LSTM model in the multi-output and online-training experiments takes 1.5 months. Finally, while the CNN model usually underperformed in most experiments under stable conditions, this is not true for the COVID-19 conditions.

To evaluate the statistical performance of the models and experiments, we regressed the ${MAAPE}_{n,t}$ with respect to the COVID-19 and protest conditions. These are dummy variables where 1 represents that the period $t$ belongs to the given condition and 0 indicates otherwise. Some temporal variables are not well captured in the models; therefore, we also controlled for Saturday and holiday effects, as follows: 

\begin{equation}
	{MAAPE}_{n,t}=\alpha_n+\ \beta_{1,n}\left({covid}_t\right)+\beta_{2,n}\left({strikes}_t\right)+\beta_{3,n}\left({temporal}_t\right)+\varepsilon_n
    \label{eq:maape}
\end{equation}

where $n$ represents the experiment, and $t$ is the aggregation period. The parameter $\alpha_n$ represents the average error of the model during stable conditions, $\beta_{1,n}$ is the added error associated with the COVID-19 pandemic, and $\beta_{2,n}$ is the added error associated with the strikes. The advantage of linear regression is the standard error estimation, which allows comparing whether the difference in the MAAPE between two models is statistically significant. The results of these estimates are provided in the appendix. 

\subsection{Stable Conditions}

\begin{figure}[h!]
\begin{center}
\includegraphics[width=0.60\textwidth]{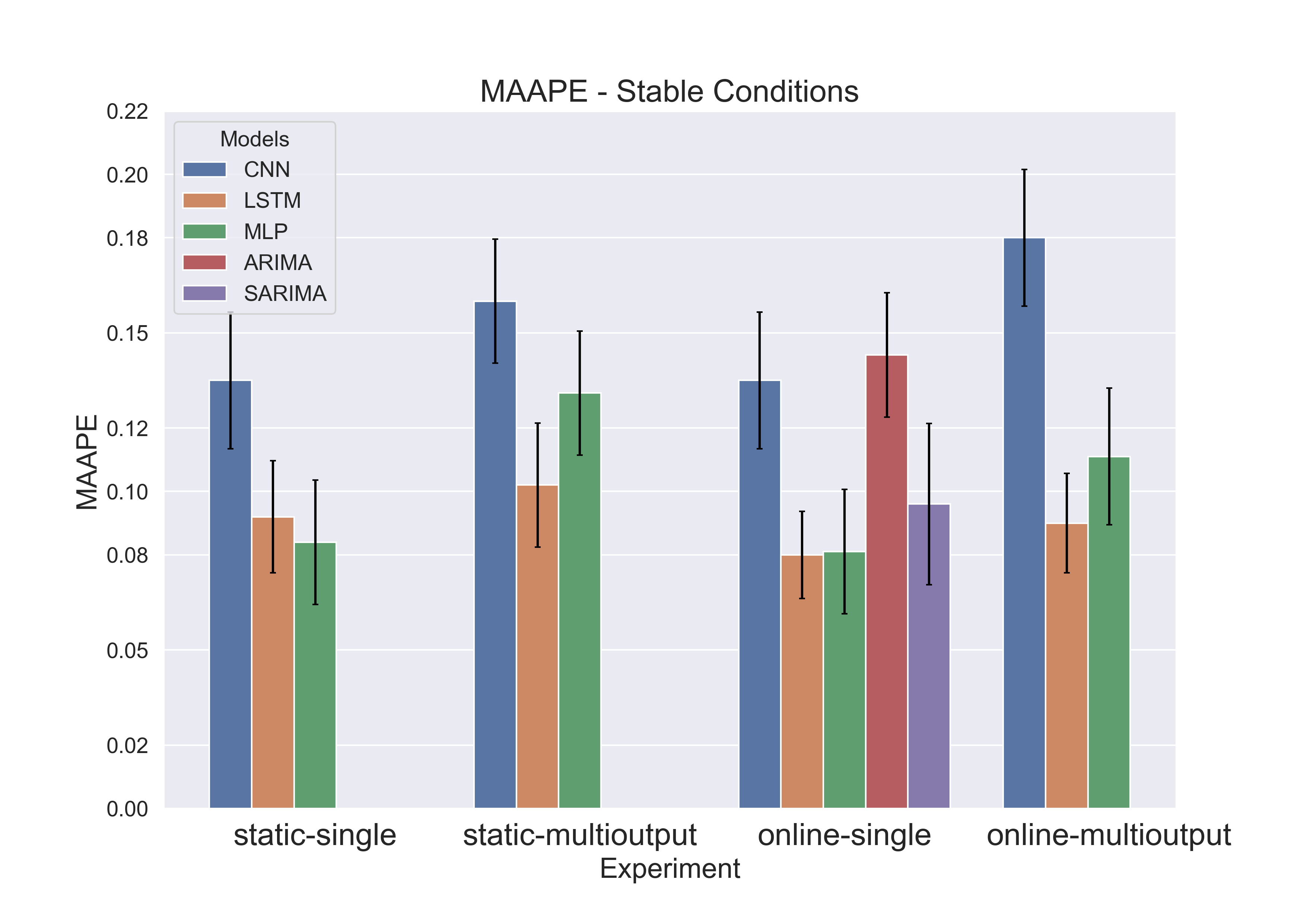}
\end{center}
\caption{Mean Arctangent Absolute Percentage Error in Stable Conditions}\label{fig-2-3:stable}
\end{figure}

For stable conditions, the single-output and online-training LSTM model performed best (Figure \ref{fig-2-3:stable}). This model significantly outperformed six of the 12 experiments. The daily MAAPE is 0.08 and its standard deviation is 0.007. Other models performed similarly, for example, the single-output and online-training dense models. However, the LSTM version is preferred because of the lower magnitude of its standard error. The multi-output version of the online-training LSTM model also performed well, with 0.09 accuracy. The multi-output with static-training CNN performed significantly worse in eight out of the 12 experiments, with a MAAPE of 0.16. 

\subsection{COVID-19 Condition}

\begin{figure}[h!]
\begin{center}
\includegraphics[width=0.60\textwidth]{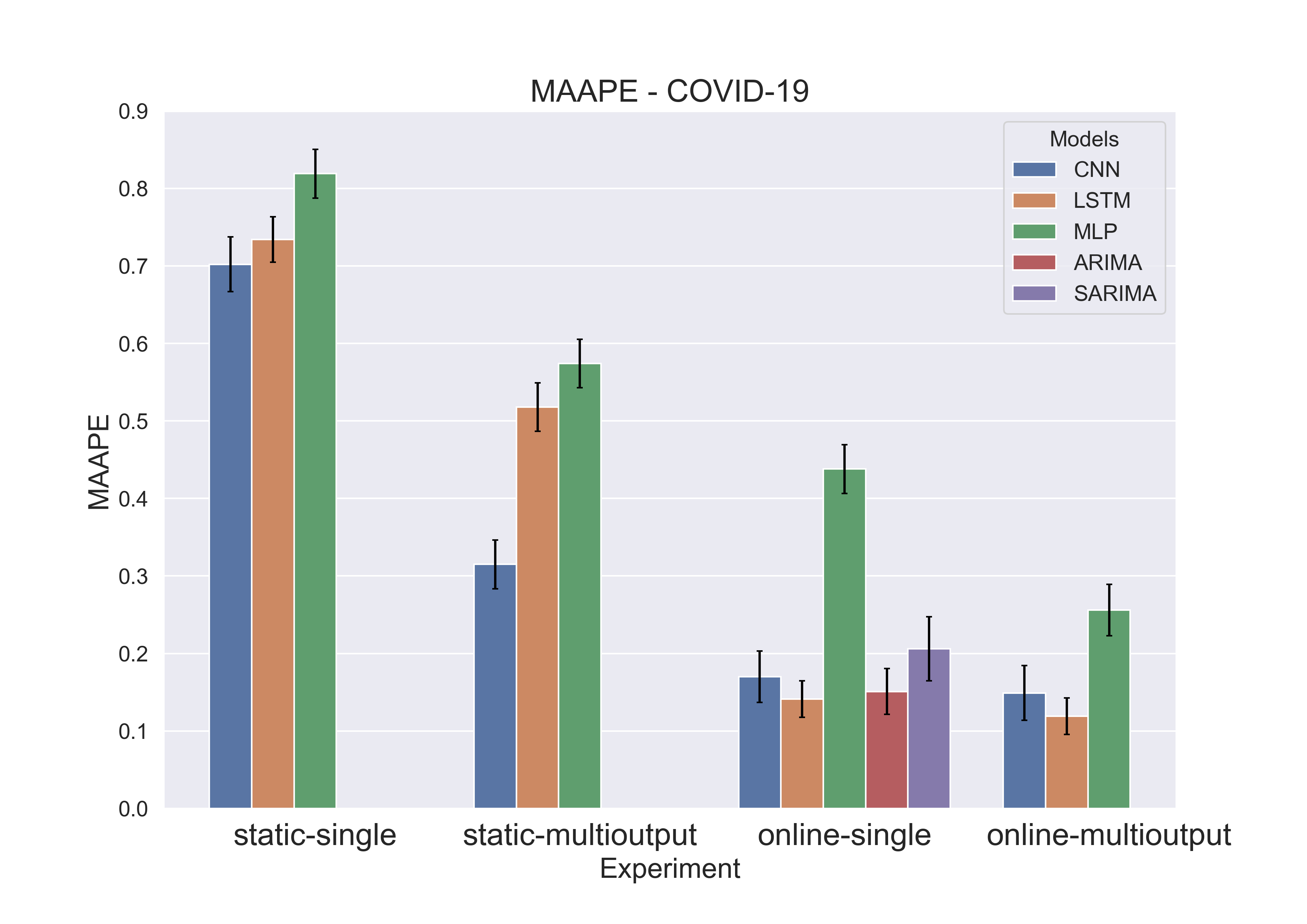}
\end{center}
\caption{Mean Arctangent Absolute Percentage Error during COVID-19 condition}\label{fig-2-3:covid}
\end{figure}

The best-performing model for the highly dynamic COVID-19 conditions was the multi-output, online-training LSTM model, as illustrated in \ref{fig-2-3:covid}. The marginal increase in the MAAPE during the COVID-19 pandemic was 0.119, with a standard error of 0.012. This model outperformed nine of the 12 experiments and has no statistically significant difference between the single-output online-training LSTM and the multi-output online-training CNN model. The worst-performing models were the single-output and static training models, with an average MAAPE increase of 0.75. As shown in Figure \ref{fig-2-2:maape_evolution}, all models have a significant increase in the MAAPE at the beginning of the COVID-19 condition, but the online-training models level up relatively quickly after a few months. 

\subsection{Protest Condition}

\begin{figure}[h!]
\begin{center}
\includegraphics[width=0.60\textwidth]{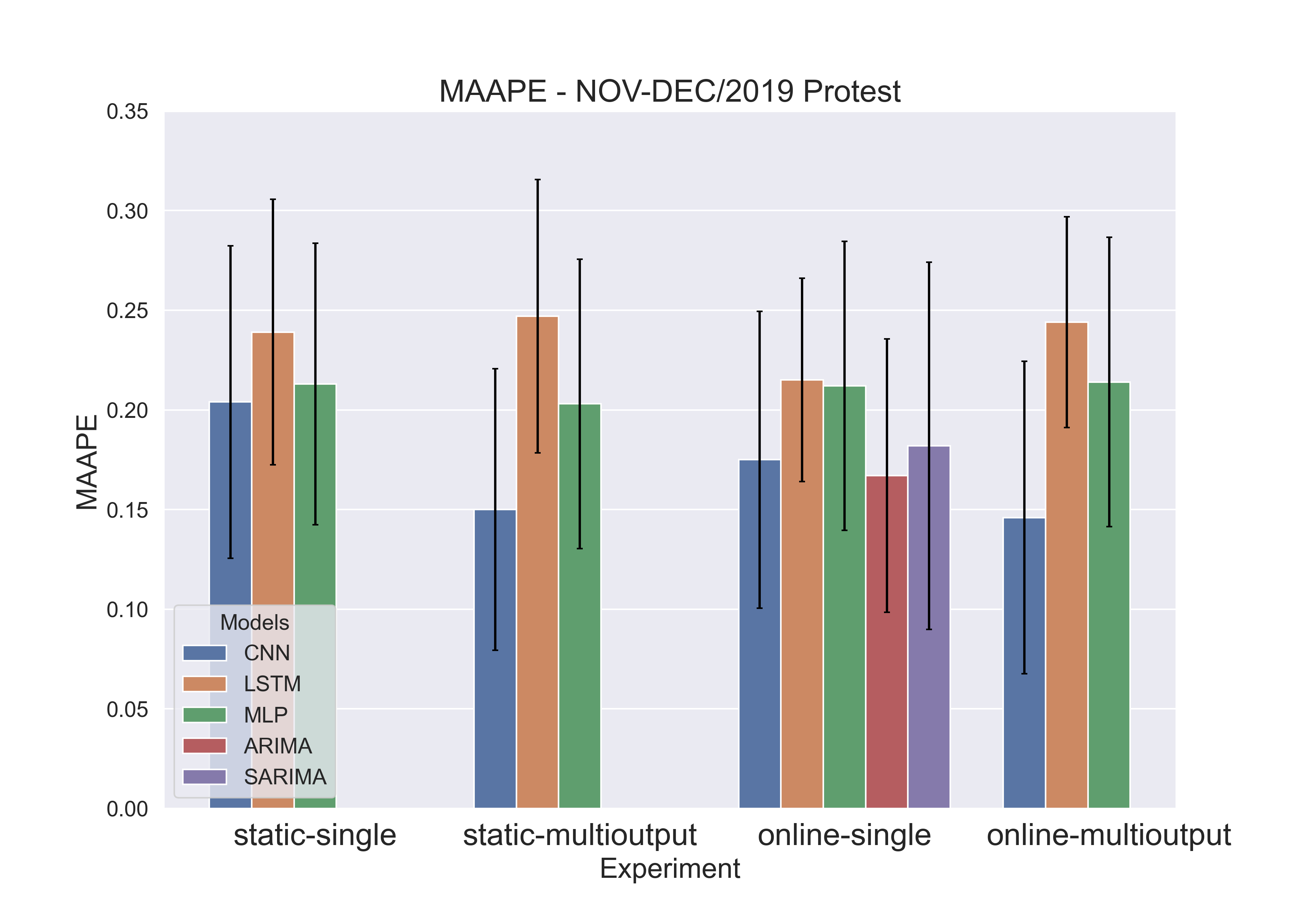}
\end{center}
\caption{Mean Arctangent Absolute Percentage Error during protest condition}\label{fig-2-3:protest}
\end{figure}

The performance during the one-month-long protest from November to December 2019 is presented in Figure \ref{fig-2-3:protest}. While the performance of any of the implementations during the month-long protest are relatively similar, the multioutput online CNN model performed slightly better. For this model, the increased MAAPE is 0.175. The multi-output, static-training LSTM model demonstrated significantly inferior performance compared to the other experiments, with an increase in MAAPE of 0.247.

\subsection{Other temporal Variables }

\begin{figure}[h]
\begin{center}
\includegraphics[width=0.60\textwidth]{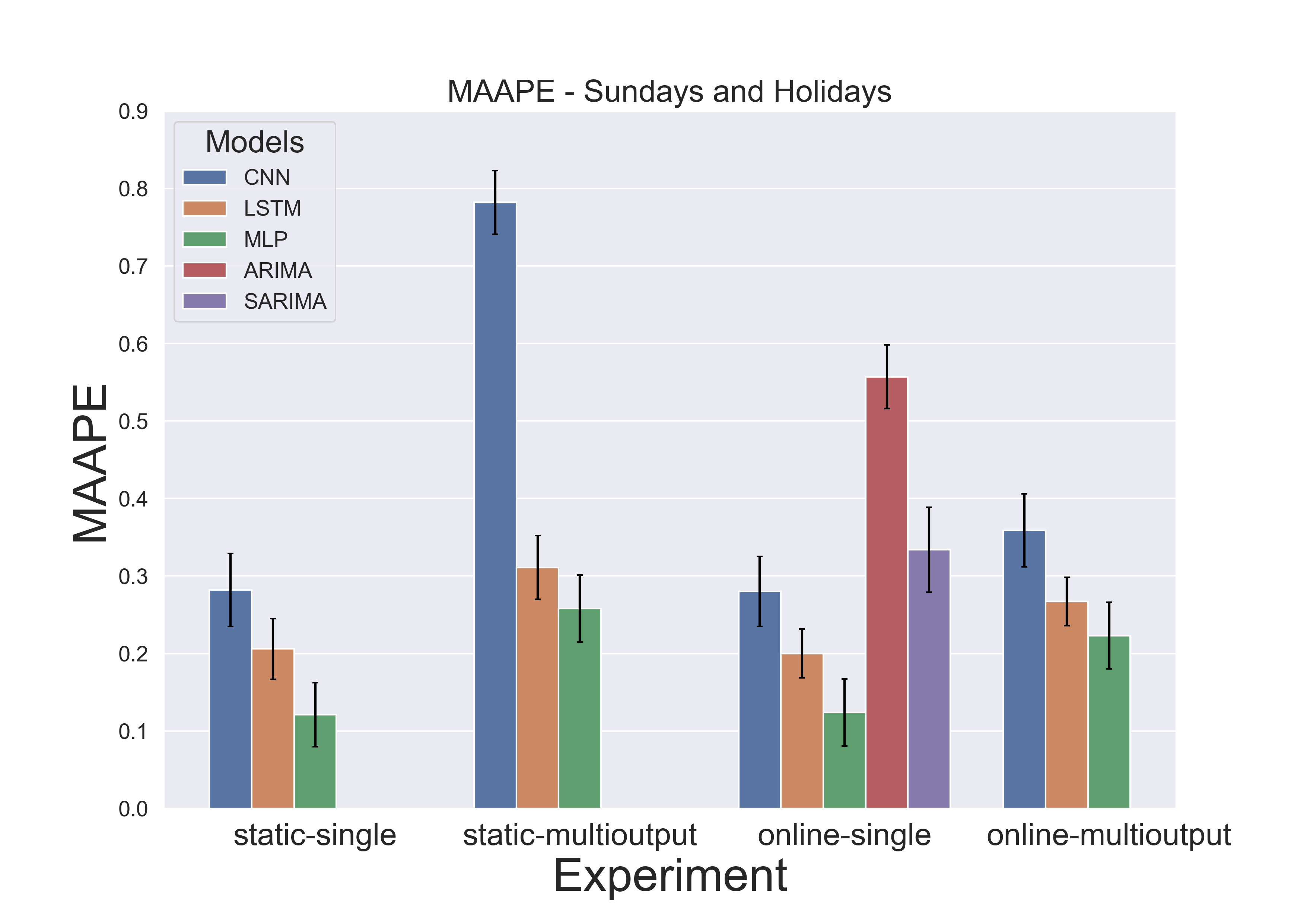}
\end{center}
\caption{Mean Arctangent Absolute Percentage Error Benchmark - Holidays}\label{fig-2-3:sundays}
\end{figure}

Figure \ref{fig-2-3:sundays} reveals that for all tested models, the MAAPE is higher during holidays, although these variables were added to the explanatory variables to account for the differentiated behavior. The single-output static-training MLP model performed best for holidays with a MAAPE of 0.121, significantly outperforming 10 out of the 12 experiments. The only model that performs similarly is its online version, with an increased MAAPE of 0.124. For holiday periods, the performance of the multi-output, static-training CNN model deteriorated, as evidenced by an increased MAAPE of 0.782. However, based on the statistical analysis, the findings indicate that the performance on Saturdays exhibits a similar level of efficacy as observed on weekdays. 

\subsection{Running times}
 
\begin{figure}[h!]
\begin{center}
\includegraphics[width=15cm]{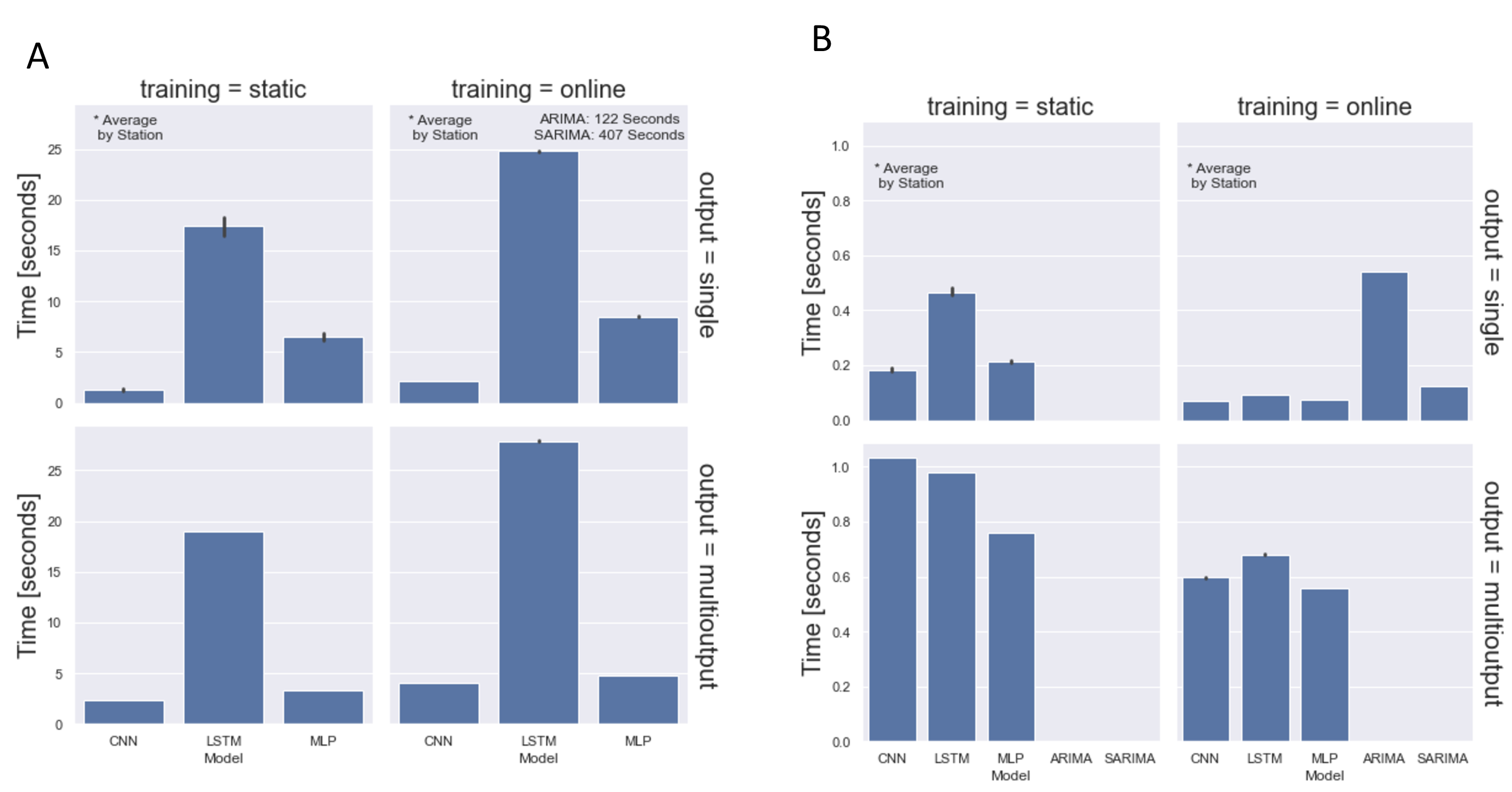}
\end{center}
\caption[Average Training and Simulation Running Times]{Average Training and Simulation Running Times A). Training B). Simulation}\label{fig-2-4:run_time}
\end{figure}

The training and simulation running time, as illustrated in Figure \ref{fig-2-4:run_time}, also serves as a significant metric for evaluation. In the case of the static models, we present the average training time for an individual station, and for the multi-output model, we provide the training time of the complete model. Therefore, to accurately compare the total training times for the entire system, single-output models must be multiplied by the number of transit stations. The difference between training a single station versus training all stations simultaneously is nearly the same, showing the superior computational efficiency of multi-output models. The training times for the ARIMA and SARIMA models were excluded from the plotted data due to a notable disparity of one order of magnitude between them. The LSTM model exhibits a significantly slower training speed compared to the other models, with a nearly five-fold difference. However, this outcome is anticipated since LSTM is an RNN architecture that inherently lacks parallelizability.

In terms of simulation times, the static models provide the simulation time for the entire test period, while the online models present the average simulation time for a single time step within the test period. Simulation running times are 1.0 s or less for all models and experiments. Multi-output models can take five times longer than single-output models, but the extra running time is offset by the fact that multi-output models predict the demand of the entire transit system at once.

In summary, the multi-output, online-training models demonstrated superior performance compared to the single-output, static-training models. Under stable conditions, both LSTM and MLP models outperformed the other models, exhibiting relatively similar performance without any statistically significant differences. Likewise, the online-training variants of these models exhibited slightly better performance than their static counterparts, although the disparity was not statistically significant. Considering the more convenient running times of multi-output models compared to single-output models, we select the multi-output, online-training LSTM model as the best-performing model for stable conditions.

In the COVID-19 condition, online models significantly outperform static models. The multi-output, online-training LSTM model emerges as the top-performing model, exhibiting an increased error of 0.12. Notably, this model demonstrates faster adaptation to new patterns compared to other models. This is evident in the sharp reduction of the MAAPE within the initial 1.5 months following the onset of the pandemic. For the long-month protest, none of the models perform significantly better than the others; however, we select the multi-output, online-training CNN model, as it performs slightly better than other models. Table \ref{tab:table3} summarizes the state-of-the-art performance for stable conditions and the two highly dynamic conditions examined in this research.

\begin{table}[h!]
	\caption{State\mbox{-}of\mbox{-}the\mbox{-}Art Performance.}
	\centering
	\begin{tabular}{llllll}
		\toprule[1.5pt]
		Condition    & Output & Training & Model & \shortstack{Training Running\\Time} & MAAPE (C.I) \\
		\midrule[1.5pt]
		Stable & Multiple & Online & LSTM & 27 sec & $0.09(\pm0.016)$  \\
        COVID-19 & Multiple & Online & LSTM & 27 sec & $0.12(\pm0.023)$  \\
        Protest & Multiple & Online & CNN & 4 sec & $0.20(\pm0.078)$  \\

		\bottomrule[1.5pt]
	\end{tabular}
	\label{tab:table3}
\end{table}

\section{Discussion and Conclusion}

Our comprehensive evaluation of various transit demand ridership prediction models highlights the pivotal role of consistent sharing, collaborating and benchmarking in advancing the field and providing actionable insights for both the research community and policymakers. This paper performed a meta-analysis to compare and critique the performance of some model architectures, including econometric and deep learning approaches, under stable and highly dynamic conditions. The dynamic conditions in this time series included the COVID-19 pandemic and a month-long protest. Both conditions triggered unexpected and uncertain closures of multiple transit stations and shifts in demand that forced changes in service operations. We demonstrated that model architectures and strategies in this analysis lost predictive power during highly dynamic conditions, as illustrated in Figure \ref{fig-2-2:maape_evolution}. The multi-output online LSTM model performed best during COVID-19 conditions, and it is also the model architecture that most rapidly adapted to highly dynamic demand. While the performance during the month-long protest was relatively similar across all tested models, the multi-output online CNN model performed slightly better.

The evidence presented in this research suggests that that online models improve accuracy faster, but learning new demand patterns takes at least 1.5 months for the tested models. This result is significant not only because the duration of a highly dynamic condition is uncertain, yet decisions to modify transit operations must be fast and prompt, but also because it exemplifies that the approach to the modeling could potentially have a more profound impact than the model architecture itself. In this research, we showed the potential of the online-training strategy to learn new patterns and improve model accuracy; however, more research is needed to learn these patterns faster. While models for stable conditions do not necessarily benefit from online training, the unexpected nature of highly dynamic conditions makes a static strategy insufficient.  In addition, more research is needed to account for station closures and their effects on the surrounding stations. The current implementations assume all stations are open, and nonzero values are predicted even when a station is closed (Appendix - Figure 8).

This research demonstrates that some implementations have similar performance, and other factors, such as running time, might be more relevant to determining a model strategy and implementation. A common benchmarking infrastructure allows researchers to readily identify shortcomings in model implementations for a given task and propose new strategies to solve them. For instance, these results demonstrate  a significant decline in the models' predictive accuracy during holidays, even when commonly used exogenous variables are added to account for these effects. Therefore, a standard and systematic benchmark process could expand faster and more reliable short-term ridership prediction.

Academic publications on short-term demand prediction have grown exponentially in the last few years, making selecting state-of-the-art models nearly impossible for researchers and practitioners. The benchmarking process in this research samples only a few models with no certainty that these implementations produce state-of-the-art performance. The establishment of a common benchmarking platform would have rendered the process of determining the most adaptable models to dynamic conditions more efficient, thus facilitating timely decision-making by policymakers. Focusing on improving model adaptability, as measured by the time taken to learn new patterns, would have provided the research community with greater focus and reliability, consequently increasing its capability to identify relevant contributions and shape decision-making. Instead, the publication process for this research extended over years, and due to the scarcity of ready-to-use model implementations, the fundamental question remains unanswered.

Therefore, we call for a reorientation of research practices in travel demand modeling, prioritizing open-source code and open data to enhance reproducibility and collaboration, rather than solely focusing on model uniqueness. Embracing open, collaborative platforms that enable systematic replicability and benchmarking is crucial for strengthening the credibility, policy-relevance and real-world impact of scientific work \citep{nosek2012s,nosek2015,camerer2016}. It allows the research community to consistently scrutinize findings, identify robust techniques, and rapidly operationalize them into practical, policy-ready solutions as new disruptions occur. Promoting transparency, validation and collective refinement equips the field to provide up-to-date decision support aligned with the shifting nature of the policy contexts it informs. An open research ecosystem that prioritizes reproducibility can help rebuild public trust, solidify science's role in evidence-based decision-making \citep{wang2022r,hendriks2020,sanders2018,uhlmann2019}, and ultimately lead to more effective, data-driven policy.

We issue an open invitation to the travel demand research community to engage with the open-source codebase and data provided herein. To remain policy-relevant in our era of disruptions, we must shift our research practices towards the principles of open science. We encourage researchers to test their own modeling approaches on this benchmarking platform, challenge the analyses conducted in this paper, and develop model specifications that can outperform those evaluated here, either using our existing metric or newly proposed ones. Our aim is to foster a collaborative, collective system where researchers contribute their methods and insights, engage in rigorous comparative testing, and jointly work towards establishing policy-relevant benchmarks and decision-support capabilities that are robust and reproducible. Through this open platform, we hope to generate collective wisdom that advances the state of the art in a field increasingly vital for effective transportation planning and policy-making as travel demand patterns continue to be disrupted.

Our open benchmarking platform provides researchers access to a centralized pool of rigorously validated models. It streamlines adaptation, implementation and comparative evaluation against consistent standards, without redundant redevelopment efforts. This collaborative hub empowers the community to collectively refine and advance transit demand modeling capabilities to remain reliable and relevant for rapid decision-making as conditions rapidly evolve.

 The open-source codebase to replicate the results of this research is available at 

\begin{center}\url{https://github.com/jdcaicedo251/transit\_demand\_prediction}
\end{center}

\textbf{Declaration of generative AI and AI-assisted technologies in the writing process}

During the preparation of this work the authors used chatGPT in order to improve language and readability. After using this tool, the authors reviewed and edited the content as needed and take full responsibility for the content of the publication.


\clearpage
\appendix




\section{Appendix}
\subsection{Statistical Analysis Results}

This appendix presents the results of the regression analysis described in Equation \ref{eq:maape}. In the equation, the intercept term represents the average MAAPE observed under stable conditions, while the remaining coefficients correspond to the increased error associated with the two highly dynamic conditions and other temporal variations. Additionally, $N$ denotes the sample size for the regression, which corresponds to the number of days in the training period. 

\begin{table}[h!]
\caption{Statistical Analysis - Single-Output and Static Training}
\centering
\begin{tabular}{lccc}
    \hline
                   & CNN       & LSTM      & MLP      \\
    \hline
    Intercept      & 0.135***  & 0.092***  & 0.084***   \\
                   & (0.011)   & (0.009)   & (0.010)    \\
    covid          & 0.702***  & 0.734***  & 0.819***   \\
                   & (0.018)   & (0.015)   & (0.016)    \\
    protest        & 0.204***  & 0.239***  & 0.213***   \\
                   & (0.040)   & (0.034)   & (0.036)    \\
    saturday       & 0.004     & 0.007     & 0.012      \\
                   & (0.026)   & (0.022)   & (0.024)    \\
    holidays       & 0.282***  & 0.206***  & 0.121***   \\
                   & (0.024)   & (0.020)   & (0.021)    \\
    covid:saturday & -0.110*** & -0.134*** & -0.214***  \\
                   & (0.042)   & (0.035)   & (0.038)    \\
    covid:holidays & -0.347*** & -0.310*** & -0.379***  \\
                   & (0.037)   & (0.031)   & (0.034)    \\
    R-squared      & 0.667     & 0.752     & 0.761      \\
    R-squared Adj. & 0.665     & 0.751     & 0.760      \\
    N              & 990  & 990  & 990   \\
    \hline
\end{tabular}
\label{tab:table_a}
\end{table}

\begin{table}[h!]
\caption{Statistical Analysis - Multi-Output and Static Training}
\label{tab:table_b}
\begin{center}
\begin{tabular}{lccc}
\hline
               & CNN       & LSTM      & MLP      \\
\hline
Intercept      & 0.160***  & 0.102***  & 0.131***   \\
               & (0.010)   & (0.010)   & (0.010)    \\
covid          & 0.315***  & 0.518***  & 0.574***   \\
               & (0.016)   & (0.016)   & (0.016)    \\
protest        & 0.150***  & 0.247***  & 0.203***   \\
               & (0.036)   & (0.035)   & (0.037)    \\
saturday       & 0.047**   & 0.019     & 0.081***   \\
               & (0.024)   & (0.023)   & (0.024)    \\
holidays       & 0.782***  & 0.311***  & 0.258***   \\
               & (0.021)   & (0.021)   & (0.022)    \\
covid:saturday & -0.053    & -0.055    & -0.330***  \\
               & (0.037)   & (0.037)   & (0.039)    \\
covid:holidays & -0.360*** & -0.136*** & -0.141***  \\
               & (0.033)   & (0.033)   & (0.034)    \\
R-squared      & 0.669     & 0.634     & 0.623      \\
R-squared Adj. & 0.667     & 0.632     & 0.621      \\
N              & 990  & 990  & 990   \\
\hline
\end{tabular}
\end{center}
\end{table}

\begin{table}[h!]
\caption{Statistical Analysis - Single-Output and Online-Training}
\label{tab:table_c}
\begin{center}
\begin{tabular}{lccccc}
\hline
               & ARIMA    & SARIMA   & CNN       & LSTM      & MLP      \\
\hline
Intercept      & 0.152*** & 0.109*** & 0.135*** & 0.080*** & 0.081***   \\
               & (0.009)  & (0.013)  & (0.011)  & (0.007)  & (0.010)    \\
covid          & 0.139*** & 0.193*** & 0.170*** & 0.141*** & 0.438***   \\
               & (0.014)  & (0.020)  & (0.017)  & (0.012)  & (0.016)    \\
protest        & 0.183*** & 0.233*** & 0.175*** & 0.215*** & 0.212***   \\
               & (0.032)  & (0.045)  & (0.038)  & (0.026)  & (0.037)    \\
saturday       & 0.022    & 0.023    & 0.007    & 0.009    & 0.013      \\
               & (0.021)  & (0.030)  & (0.025)  & (0.017)  & (0.024)    \\
holidays       & 0.547*** & 0.348*** & 0.280*** & 0.200*** & 0.124***   \\
               & (0.019)  & (0.027)  & (0.023)  & (0.016)  & (0.022)    \\
covid:saturday & 0.059*   & -0.037   & 0.021    & 0.065**  & -0.139***  \\
               & (0.034)  & (0.048)  & (0.040)  & (0.028)  & (0.039)    \\
covid:holidays & 0.045    & -0.009   & -0.019   & 0.045*   & -0.110***  \\
               & (0.030)  & (0.042)  & (0.036)  & (0.025)  & (0.034)    \\
R-squared      & 0.630    & 0.302    & 0.289    & 0.398    & 0.487      \\
R-squared Adj. & 0.628    & 0.297    & 0.285    & 0.394    & 0.484      \\
N              & 990 & 990 & 990 & 990 & 990   \\
\hline
\end{tabular}
\end{center}
\end{table}

\begin{table}[h!]
\caption{Statistical Analysis - Multi-Output and Online-Training}
\label{tab:table_d}
\begin{center}
\begin{tabular}{lccc}
\hline
               & CNN       & LSTM      & MLP     \\
\hline
Intercept      & 0.180*** & 0.090*** & 0.111***  \\
               & (0.011)  & (0.008)  & (0.011)   \\
covid          & 0.149*** & 0.119*** & 0.256***  \\
               & (0.018)  & (0.012)  & (0.017)   \\
protest        & 0.146*** & 0.244*** & 0.214***  \\
               & (0.040)  & (0.027)  & (0.037)   \\
saturday       & -0.010   & 0.014    & 0.023     \\
               & (0.026)  & (0.018)  & (0.025)   \\
holidays       & 0.359*** & 0.267*** & 0.223***  \\
               & (0.024)  & (0.016)  & (0.022)   \\
covid:saturday & 0.053    & 0.065**  & 0.029     \\
               & (0.042)  & (0.028)  & (0.040)   \\
covid:holidays & 0.064*   & 0.001    & 0.191***  \\
               & (0.037)  & (0.025)  & (0.035)   \\
R-squared      & 0.379    & 0.412    & 0.453     \\
R-squared Adj. & 0.375    & 0.409    & 0.450     \\
N              & 990 & 990 & 990  \\
\hline
\end{tabular}
\end{center}
\end{table}

\clearpage

\subsection{Closed Stations Analysis}

\begin{figure}[h!]
\centering
\includegraphics[width=0.65\textwidth]{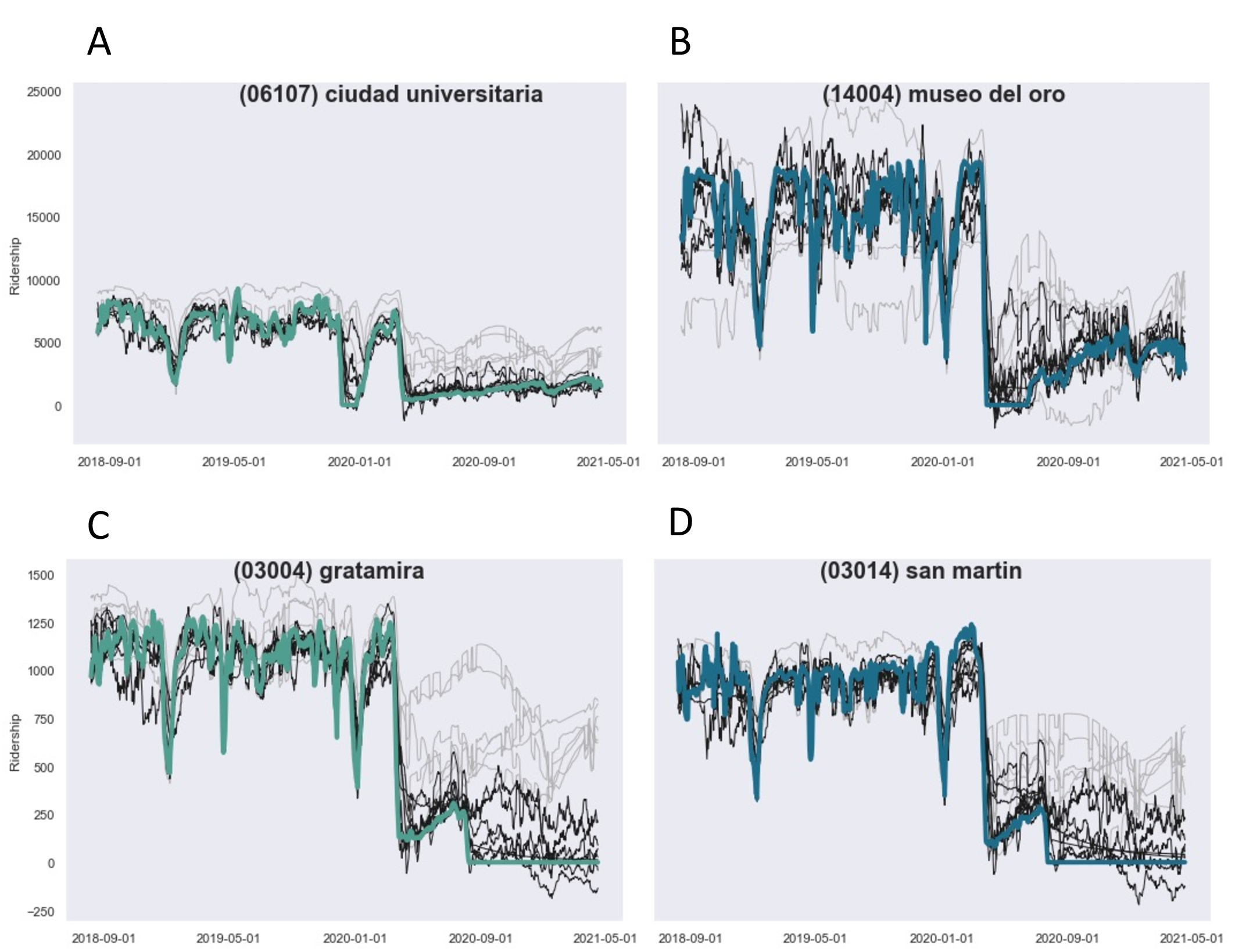}
\caption[Temporary Closed Stations examples]{Temporary Closed Stations examples. Light-gray: Static Models. Dark-gray: Online Models}\label{fig:figd}
\end{figure}

\end{document}